\documentclass[10pt,twocolumn,letterpaper]{article}

\usepackage{cvpr}
\usepackage{times}
\usepackage{epsfig}
\usepackage{graphicx}
\usepackage{amsmath}
\usepackage{amssymb}

\usepackage{subfig}
\usepackage{multirow}
\usepackage[bookmarks=true,bookmarksnumbered=true,
bookmarksopen=false,colorlinks=true]{hyperref} 

\cvprfinalcopy 

\begin{document}

\title{HMRF-EM-image: Implementation of the Hidden Markov Random Field Model 
and its Expectation-Maximization Algorithm}

\author{Quan Wang\\
Signal Analysis and Machine Perception Laboratory\\
Electrical, Computer, and Systems Engineering\\
Rensselaer Polytechnic Institute\\
{\tt\small wangq10@rpi.edu}
}

\maketitle

\begin{abstract}
In this project\footnote{This work originally appears as the final project of 
Prof. \href{http://www.ecse.rpi.edu/~yazici/}{Birsen Yazici}'s course \emph{Detection and Estimation Theory} at RPI.}, 
we study the hidden Markov random field (HMRF) model and its expectation-maximization (EM) algorithm. We implement a {\tt \textit{MATLAB}} toolbox named {\tt \textit{HMRF-EM-image}} for 2D image segmentation using the HMRF-EM framework\footnote{This toolbox can be downloaded at the author's homepage 
\href{http://homepages.rpi.edu/~wangq10}{http://homepages.rpi.edu/~wangq10}.}. 
This toolbox also implements 
edge-prior-preserving image segmentation, and can be easily reconfigured for other problems, 
such as 3D image segmentation. 
\end{abstract}

\section{Introduction}
Markov random fields (MRFs) have been widely used for computer vision problems, such as
image segmentation \cite{zhanglei}, surface reconstruction \cite{surface} 
and depth inference \cite{depth}. Much of its success attributes to the efficient algorithms, 
such as Iterated Conditional Modes \cite{ICM}, and its consideration of both 
``data faithfulness'' and ``model smoothness'' \cite{AGSM}. 

The HMRF-EM framework was first proposed for segmentation of brain MR images
\cite{HMRF-EM}. 
Given an image $\textbf{y}=(y_1,\dots,y_N)$ where each $y_i$ is the intensity of a pixel, 
we want to infer a configuration of labels $\textbf{x}=(x_1,\dots,x_N)$ where $x_i \in L$ and $L$
is the set of all possible labels. In a binary segmentation
problem, $L= \lbrace 0,1 \rbrace$. According to the MAP 
criterion, we seek the labeling $\textbf{x}^\star$ which satisfies 
\begin{equation}
\label{eq:MAP}
\textbf{x}^\star=\underset{\textbf{x}}{\operatorname{argmax}} \; \lbrace 
P(\textbf{y}|\textbf{x},\Theta)P(\textbf{x}) \rbrace  . 
\end{equation}
The prior probability $P(\textbf{x})$ is a Gibbs distribution, and the joint likelihood
probability is 
\begin{eqnarray}
P(\textbf{y}|\textbf{x},\Theta)&=&\prod\limits_i P(y_i|\textbf{x},\Theta) \nonumber \\ 
&=&\prod\limits_i P(y_i|x_i,\theta_{x_i})  , 
\end{eqnarray}
where $P(y_i|x_i,\theta_{x_i})$ is a Gaussian distribution with parameters
$\theta_{x_i}=(\mu_{x_i},\sigma_{x_i})$.
$\Theta=\lbrace \theta_l | l \in L \rbrace$ is the parameter set, which is obtained
by the EM algorithm. 

\section{EM Algorithm}
We use the EM algorithm to estimate the parameter set 
$\Theta=\lbrace \theta_l | l \in L \rbrace$. We describe the EM algorithm by the following:
\begin{enumerate}
\item
\textit{Start:} Assume we have an initial parameter set $\Theta^{(0)}$.

\item
\textit{E-step:} At the $t$th iteration, we have $\Theta^{(t)}$, and 
we calculate the conditional expectation:
\begin{eqnarray}
Q(\Theta | \Theta^{(t)}) &=& E\left[ 
\ln P(\textbf{x},\textbf{y}|\Theta) | \textbf{y},\Theta^{(t)}
\right]
\nonumber \\
&=& \sum\limits_{\textbf{x}\in \chi}
P(\textbf{x}|\textbf{y},\Theta^{(t)}) \ln P(\textbf{x},\textbf{y}|\Theta)  , 
\end{eqnarray}
where $\chi$ is the set of all possible configurations of labels. 

\item
\textit{M-step:} Now maximize $Q(\Theta | \Theta^{(t)})$
to obtain the next estimate:
\begin{equation}
\Theta^{(t+1)}=\underset{\Theta}{\operatorname{argmax}} \;  Q(\Theta | \Theta^{(t)}) . 
\end{equation}
Then let $\Theta^{(t+1)} \rightarrow \Theta^{(t)}$ and repeat from the E-step. 
\end{enumerate}

Let $G(z;\theta_l)$ denote a Gaussian distribution function with parameters 
$\theta_l=(\mu_l,\sigma_l)$:
\begin{equation}
G(z;\theta_l)=\dfrac{1}{\sqrt{2\pi\sigma_l^2}} \exp\left(
-\dfrac{(z-\mu_l)^2}{2\sigma_l^2}
\right) . 
\end{equation}
We assume that the prior probability can be written as 
\begin{eqnarray}
P(\textbf{x})=\dfrac{1}{Z}\exp\left( -U(\textbf{x}) \right) , 
\end{eqnarray}
where $U(\textbf{x})$ is the prior energy function. 
We also assume that 
\begin{eqnarray}
P(\textbf{y}|\textbf{x},\Theta)&=&
\prod\limits_i P(y_i|x_i,\theta_{x_i}) \nonumber \\
&=& \prod\limits_i G(y_i;\theta_{x_i}) \nonumber \\
&=& \dfrac{1}{Z'}\exp\left( -U(\textbf{y}|\textbf{x}) \right) . 
\end{eqnarray}
With these assumptions, the HMRF-EM algorithm is given below:
\begin{enumerate}
\item 
Start with initial parameter set $\Theta^{(0)}$.

\item
Calculate the likelihood distribution $P^{(t)}(y_i|x_i,\theta_{x_i})$.

\item
Using current parameter set $\Theta^{(t)}$ to estimate the labels by
MAP estimation:
\begin{eqnarray}
\textbf{x}^{(t)}&=&\underset{\textbf{x}\in \chi}{\operatorname{argmax}} \; \lbrace 
P(\textbf{y}|\textbf{x},\Theta^{(t)})P(\textbf{x}) \rbrace \nonumber \\
&=&\underset{\textbf{x}\in \chi}{\operatorname{argmin}} \; \lbrace 
U(\textbf{y}|\textbf{x},\Theta^{(t)})+U(\textbf{x}) \rbrace  . 
\end{eqnarray}
The algorithm for the MAP estimation is discussed in Section \ref{section:MAP}. 

\item
Calculate the posterior distribution for all $l\in L$ and all pixels $y_i$:
\begin{eqnarray}
P^{(t)}(l | y_i)=\dfrac{G(y_i;\theta_l)P(l | x_{N_i}^{(t)})}{P^{(t)}(y_i)} , 
\end{eqnarray}
where $x_{N_i}^{(t)}$ is the neighborhood configuration of $x_i^{(t)}$, and 
\begin{equation}
P^{(t)}(y_i)=\sum\limits_{l\in L}G(y_i;\theta_l)P(l | x_{N_i}^{(t)}) . 
\end{equation}
Note here we have 
\begin{eqnarray}
P(l | x_{N_i}^{(t)}) &=& 
\dfrac{1}{Z}\exp \left( -\sum\limits_{j\in N_i} V_c(l,x_j^{(t)})\right) . 
\end{eqnarray}

\item
Use $P^{(t)}(l | y_i)$ to update the parameters:
\begin{eqnarray}
\mu_l^{(t+1)}&=&\dfrac{\sum\limits_{i}P^{(t)}(l|y_i)y_i}{\sum\limits_{i}P^{(t)}(l|y_i)}  , \\
(\sigma_l^{(t+1)})^2&=&
\dfrac{\sum\limits_{i}P^{(t)}(l|y_i)(y_i-\mu_l^{(t+1)})^2}{\sum\limits_{i}P^{(t)}(l|y_i)}  . 
\end{eqnarray}
\end{enumerate}

\section{MAP Estimation}
\label{section:MAP}
In the EM algorithm, we need to solve for $\textbf{x}^\star$ that minimizes the total 
posterior energy 
\begin{equation}
\label{eq:MAP2}
\textbf{x}^{\star}
=\underset{\textbf{x}\in \chi}{\operatorname{argmin}} \; \lbrace 
U(\textbf{y}|\textbf{x},\Theta)+U(\textbf{x}) \rbrace 
\end{equation}
with given $\textbf{y}$ and $\Theta$, where the likelihood energy is 
\begin{eqnarray}
U(\textbf{y}|\textbf{x},\Theta)&=&
\sum\limits_{i} U(y_i|x_i,\Theta) \nonumber \\
&=& \sum\limits_{i} \left[
\dfrac{(y_i-\mu_{x_i})^2}{2\sigma_{x_i}^2}+\ln \sigma_{x_i} \right] . 
\end{eqnarray}
The prior energy function $U(\textbf{x})$ has the form 
\begin{equation}
U(\textbf{x})=\sum\limits_{c\in C} V_c(\textbf{x}) , 
\end{equation}
where $V_c(\textbf{x})$ is the clique potential and $C$ is the set of all possible cliques. 

In the image domain, we assume that one pixel has at most 4 neighbors: the pixels in 
its 4-neighborhood. Then the clique potential is defined on pairs of neighboring pixels:
\begin{eqnarray}
V_c(x_i,x_j)=\dfrac{1}{2}(1-I_{x_i,x_j}) , 
\end{eqnarray}
where 
\begin{equation}
I_{x_i,x_j}=
\left\{\begin{array}{c}
0 \qquad \textrm{if $x_i \neq x_j$}\\ 
1 \qquad \textrm{if $x_i = x_j$}
\end{array}\right. . 
\end{equation}

We have developed an iterative algorithm to solve (\ref{eq:MAP2}):
\begin{enumerate}
\item
To start with, we have an initial estimate $\textbf{x}^{(0)}$, which is from the
previous loop of the EM algorithm. 

\item 
\label{item:MAP_iter}
Provided $\textbf{x}^{(k)}$, for all $1\leq i\leq N$, we find 
\begin{equation}
\label{eq:MAP_iter}
x_i^{(k+1)}=
\underset{l \in L}{\operatorname{argmin}} \; \lbrace
U(y_i | l)+\sum\limits_{j\in N_i} V_c(l,x_j^{(k)})
\rbrace  . 
\end{equation}

\item
Repeat step \ref{item:MAP_iter} until $U(\textbf{y}|\textbf{x},\Theta)+U(\textbf{x})$ converges 
or a maximum $k$ is achieved. 
\end{enumerate}

\begin{table*}
\begin{center}
\begin{tabular}{|c|c|c|}
\hline
File & Type &
Usage
\\ \hline
\multirow{2}{*}{{\tt demo.m}} & \multirow{2}{*}{Runnable script} &
A demo showing how to use the toolbox. \\ & & Users can run this file directly. \\ \hline
\multirow{2}{*}{{\tt image\_kmeans.m}} & \multirow{2}{*}{Function} &
The k-means algorithm for 2D images. \\ & &
This will generate an initial segmentation. \\ \hline
{\tt HMRF\_EM.m} & Function & The HMRF-EM algorithm. \\ \hline
{\tt MRF\_MAP.m} & Function & The MAP algorithm. \\ \hline
{\tt gaussianBlur.m} & Function & Blurring an image using Gaussian kernel. \\ \hline
{\tt gaussianMask.m} & Function & Obtaining the mask of Gaussian kernel. \\ \hline
{\tt ind2ij.m} & Function & Index to 2D image coordinates conversion. \\ \hline
{\tt BoundMirrorExpand.m} & Function & Expanding an image. \\ \hline
{\tt BoundMirrorShrink.m} & Function & Shrinking an image. \\ \hline
{\tt Beijing World Park 8.JPG} & Image & An example input image.
\\ \hline
\end{tabular}
\end{center}
\caption{Name and usage of each file in the {\tt \textit{HMRF-EM-image}} toolbox. }
\label{table:toolbox}
\end{table*}

\section{Edge-Prior-Preserving Image Segmentation}
To use HMRF-EM framework for image segmentation, first we generate an 
initial segmentation using k-means clustering on the gray-level intensities of pixels. 
The initial segmentation provides the initial labels $\textbf{x}^{(0)}$ for the MAP algorithm, and
the initial parameters $\Theta^{(0)}$ for the EM algorithm. Then we run the EM algorithm, and 
the resulting label configuration $\textbf{x}$ will be a refined segmentation result. 

Now we would like our segmentation to preserve the edges obtained by 
some edge detection algorithm, such as Canny edge detection \cite{Canny}, Sarkar-Boyer edge detection \cite{Sarkar}, or Berkeley's contour detection \cite{berkeley}. 
Assume we have a binary edge map $\textbf{z}$, where $z_i=1$ if the $i$th pixel is on an edge, 
and $z_i=0$ if not. Then we modify (\ref{eq:MAP_iter}) to
\begin{equation}
x_i^{(k+1)}=
\underset{l \in L}{\operatorname{argmin}} \; \lbrace
U(y_i | l)+\sum\limits_{j\in N_i, z_j=0} V_c(l,x_j^{(k)})
\rbrace . 
\end{equation}

\begin{figure*}
  \centering
  \subfloat[]
  {\includegraphics[width=0.27\textwidth]{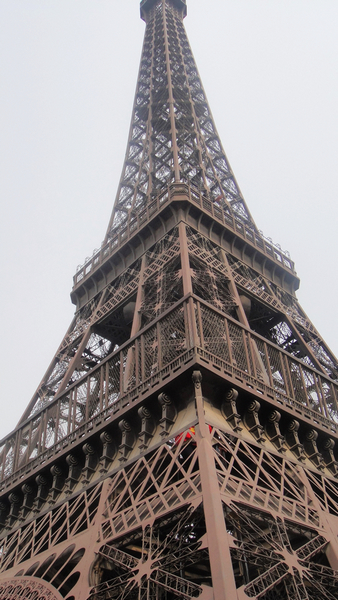} }  
  \quad      
  \subfloat[]
  {\includegraphics[width=0.27\textwidth]{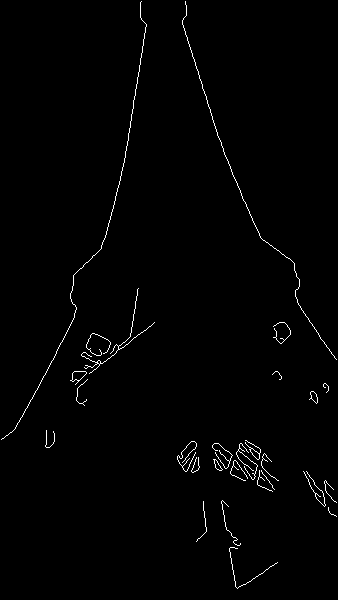} }
  \quad      
  \subfloat[]
  {\includegraphics[width=0.27\textwidth]{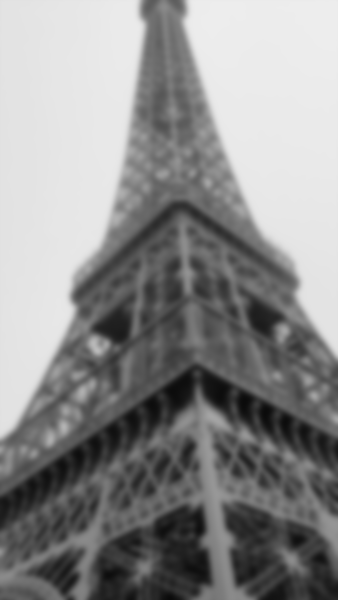} }
   \quad      
  \subfloat[]
  {\includegraphics[width=0.27\textwidth]{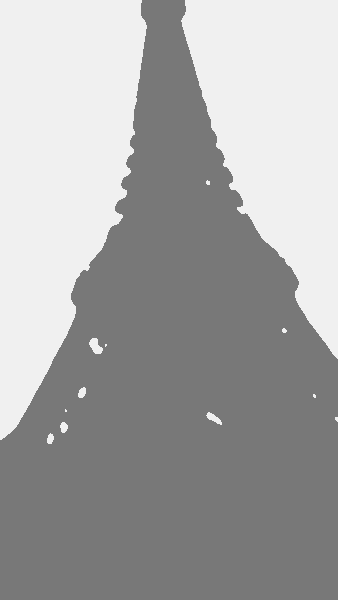} }
   \quad      
  \subfloat[]
  {\includegraphics[width=0.27\textwidth]{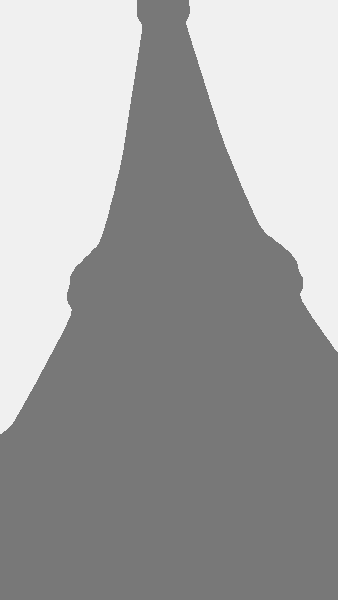} }
   \quad      
  \subfloat[]
  {\includegraphics[width=0.4\textwidth]{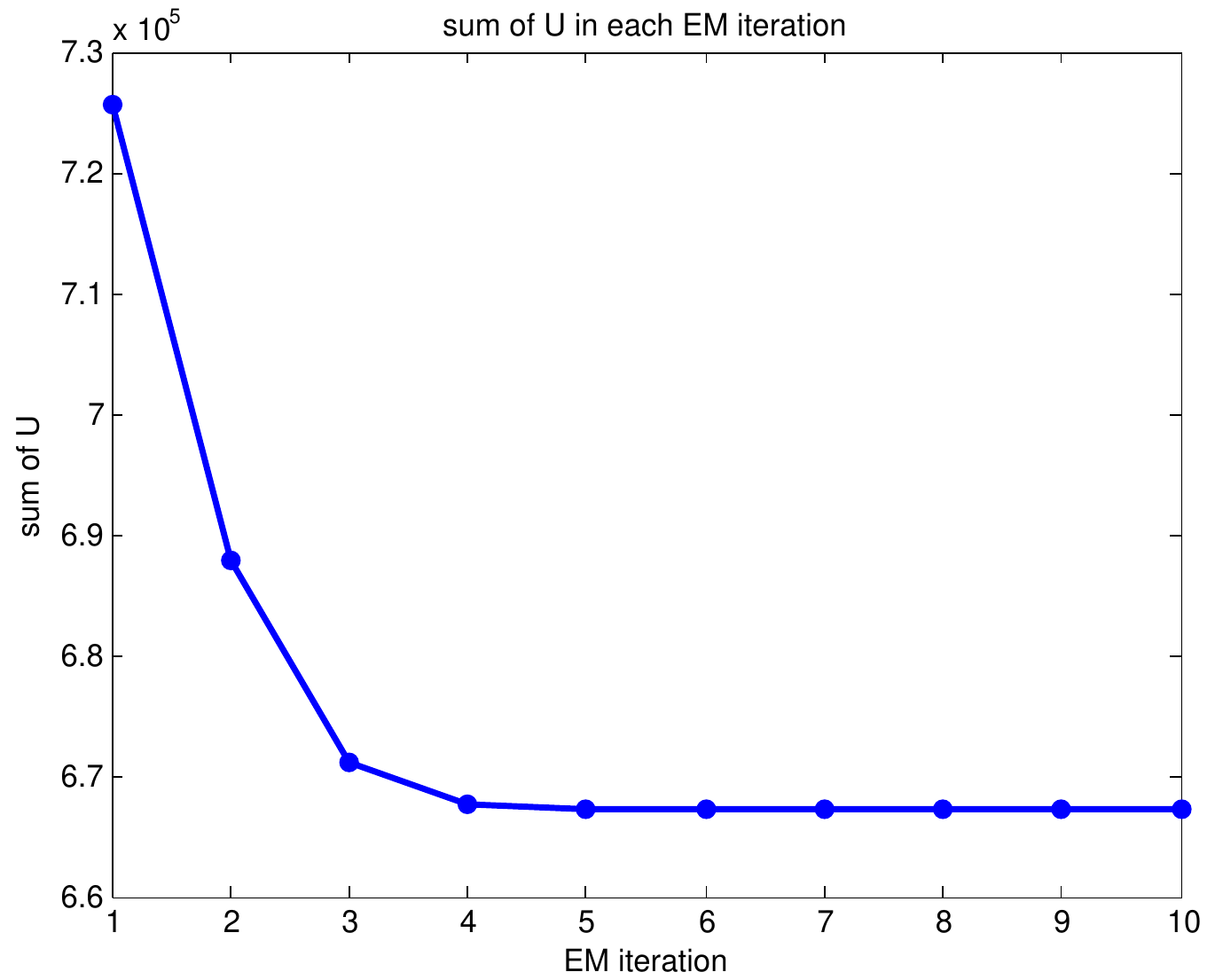} }
  \caption{Edge-prior-preserving image segmentation results. 
  (a) Original image. 
  (b) Canny edges.
  (c) Gaussian blurred image. 
  (d) Initial labels obtained by k-means, where $k=2$.
  (e) Final labels obtained by HMRF-EM algorithm. 
  (f) Total posterior energy in each iteration of the EM algorithm. 
  }
  \label{fig:tower}
\end{figure*}

\section{Experiment Results}
We run our HMRF-EM edge-prior-preserving segmentation algorithm on example images. The
binary edge map $\textbf{z}$ is obtained by performing Canny edge detection \cite{Canny} on the original image, and the observation $\textbf{y}$ is obtained by performing Gaussian blur on
the original image. Some results are shown in Figure \ref{fig:tower}. We can see that the initial labels
obtained by the k-means algorithm are not smooth enough, have morphological holes, and do not preserve the 
Canny edges. The HMRF refined labels overcome all these disadvantages. With 10 EM iterations and 
10 MAP iterations, the segmentation of a $600 \times 338$ image takes about 
40 seconds on a 2.53GHz Intel(R) Core(TM) i5 CPU.

\section{Toolbox Documentation}
We provide the name and usage of each file in our  {\tt MATLAB} toolbox
{\tt HMRF-EM-image} in Tabel \ref{table:toolbox}. 
The {\tt U\_X.m} file can be modified to re-define pixel neighborhood relationships, 
and the {\tt U\_l.m} file can be modified to re-define the clique potentials. 
To reconfigure this toolbox for 3D image segmentation, the indexing system must be modified in
several files.

\section{Discussion}
Our {\tt HMRF-EM-image} toolbox is an implementation of the hidden Markov random field
and its EM algorithm. 
This toolbox is well commented and easy to reconfigure.
We have demonstrated  the effectiveness of our toolbox on a simple example image.
The HMRF model is mainly used to refine the direct
segmentation output of some other algorithms. 
To get better segmentation results on more complicated
images, some higher-level features should be used to construct $\textbf{y}$
instead of just pixel intensities, and some more advanced algorithm should be
used to generate the initial labels. 

{\small
\bibliographystyle{ieee}
\bibliography{egbib}
}

\end{document}